\pdfoutput=1

\documentclass[11pt]{article}

\usepackage[]{acl}

\usepackage{times}
\usepackage{latexsym}

\usepackage[T1]{fontenc}

\usepackage[utf8]{inputenc}

\usepackage{microtype}
\usepackage{graphicx}
\usepackage{tabu}
\usepackage{multirow}

\newcommand{\cut}[1]{}

\newcommand{\para}[1]{{\noindent\textbf{#1}}}

\title{An Empirical Investigation of Commonsense Self-Supervision with Knowledge Graphs}

\author{
Jiarui Zhang\textsuperscript{\rm 1,2},
Filip Ilievski\textsuperscript{\rm 2},
Kaixin Ma\textsuperscript{\rm 3},
Jonathan Francis\textsuperscript{\rm 3,4} \and Alessandro Oltramari\textsuperscript{\rm 4} \\
\textsuperscript{\rm 1}Department of Electronic Engineering, Tsinghua University\\
\textsuperscript{\rm 2}Information Sciences Institute, Viterbi School of Engineering, University of Southern California\\
\textsuperscript{\rm 3}Language Technologies Institute, School of Computer Science, Carnegie Mellon University\\
\textsuperscript{\rm 4}Human-Machine Collaboration, Bosch Research Pittsburgh
}

\begin{document}
\maketitle
\begin{abstract}
Self-supervision based on the information extracted from large knowledge graphs has been shown to improve the generalization of language models, in zero-shot evaluation on various downstream language reasoning tasks. Since these improvements are reported in aggregate, however, little is known about (i) how to select the appropriate knowledge for solid performance across tasks, (ii) how to combine this knowledge with neural language models, and (iii) how these pairings affect granular task performance. In this paper, we study the effect of knowledge sampling strategies and sizes that can be used to generate synthetic data for adapting language models. We study the effect of different synthetic datasets on language models with various architectures and sizes. The resulting models are evaluated against four task properties: domain overlap, answer similarity, vocabulary overlap, and answer length. Our experiments show that encoder-decoder models benefit from more data to learn from, whereas sampling strategies that balance across different aspects yield best performance. Most of the improvement occurs on questions with short answers and dissimilar answer candidates, which corresponds to the characteristics of the data used for pre-training.
\end{abstract}

\section{Introduction}
\label{sec:intro}

Common sense is the human knowledge about the world and the methods for making inferences from this knowledge~\cite{davis2014representations}. Commonsense knowledge includes the basic facts about events (including actions) and their effects, facts about knowledge and how it is obtained, facts about beliefs and desires, as well as the basic facts about material objects and their properties~\cite{mccarthy1989artificial}. Artificial Intelligence (AI) agents that are equipped with common sense are expected to possess a wide range of everyday knowledge about naive physics, folk psychology, and causality. Rich commonsense knowledge can be found in public knowledge graphs (KGs), like ConceptNet~\cite{10.5555/3298023.3298212}, ATOMIC~\cite{DBLP:conf/aaai/SapBABLRRSC19}, and Visual Genome~\cite{krishna2017visual}.

State-of-the-art commonsense reasoning systems are largely fueled by language models (LMs), as LMs are able to adapt (\textit{fine-tune}) to 
benchmarks effectively, insofar as training data is available~\cite{devlin2018bert,liu2019roberta}.
Recent work has proposed lightweight alternatives to fine-tuning, such as Prefix-tuning~\cite{li2021prefixtuning} and AutoPrompt~\cite{autoprompt:emnlp20}, however, these methods still rely on training data being available and have limited generalizability to other tasks~\cite{ma2021exploring}, given that the data distribution varies greatly across tasks. Recognizing that the assumption of always having benchmark-specific training data is unrealistic for open-domain reasoning, recent work has increasingly focused on zero- and few-shot tasks and reasoning models. Common methods for zero-shot reasoning rely on careful pre-training of LMs with external resources: commonsense KGs~\cite{Banerjee2020SelfsupervisedKT, Ma2021}, elicitation of pre-existing knowledge in the LM~\cite{Shwartz2020UnsupervisedCQ,paranjape-etal-2021-prompting}, or instruction-prompted training with a diverse set of tasks \cite{sanh2021multitask}. While pre-training with commonsense knowledge has been shown to improve model performance~\cite{Banerjee2020SelfsupervisedKT,Ma2021}, prior work has not investigated how different architectural and data decisions affect model accuracy and generalization across tasks.



This paper conducts an empirical investigation of commonsense self-supervision of state-of-the-art language models with knowledge graphs. We study the interplay between methods for data generation from knowledge graphs, choices of language models, and task properties. Our contributions are as follows:
\begin{enumerate}
    \item We formalize the self-supervision of language models with synthetic data extracted from knowledge graphs, as a joint framework. We consider the interplay between the selected knowledge as synthetic data, the language model, and the properties of the task.
    \item We pose five research questions, which have not been answered so far for commonsense QA models adapted with KGs under zero-shot evaluation setting. We study the following aspects: (i) knowledge sampling size and strategy; (ii) language model architecture and size; and (iii) task properties: domain overlap, vocabulary overlap, answer similarity, and answer length.
    \item We gain insight into these research questions, through rigorous experimentation: we evaluate seven sampling strategies with seven knowledge sizes, which are used to adapt five models belonging to two representative model architectural classes. We define fourteen different task partitions, for five different benchmarks. Our experiments show that: (i) the best sampling strategy balances across different properties of the data, (ii) the optimal amount of data is model-dependent, (iii) most of the improvement occurs on questions with short answers and dissimilar answer candidates.
    
\end{enumerate}
\section{Research Questions}
\label{sec:hypotheses}



We study five questions which have not been fully answered by prior work on zero-shot question answering with knowledge graphs. We motivate each question and indicate the novelty introduced by studying the question in our setting.

\noindent \textit{RQ1: What is the overall impact of model and knowledge choices on the generalizability of self-supervision of LMs with KGs?} Prior work on zero-shot commonsense reasoning with KGs~\cite{Ma2021,dou2022zero} has reported large gains across benchmarks, over LM baselines. Yet, the gap between these results and the performance of supervised models remains large. It is unclear how much this gap can be bridged by tuning the knowledge and model selection in the self-supervision method.


\noindent \textit{RQ2: How much data is needed to adapt LMs to commonsense reasoning tasks?} Finding a right number of QA pairs to adapt a model with is crucial to reach optimal performance, prevent overfitting, and optimize efficiency. Ma et al.~\cite{Ma2021} report accuracy gains with a hand-selected subset of the CSKG graph, whereas Ilievski et al.~\cite{ilievski2021dimensions} show that adapting language models with questions from certain knowledge dimensions is much more beneficial than others, and sometimes, even better than using the entire set of questions. No prior work has performed systematic analysis of the relation between knowledge sample size and the model performance.

\noindent \textit{RQ3: How to best sample questions for model adaptation?} Even after detecting the optimal data size, it remains unclear how to best select the needed data points. For instance, sampling can focus on optimizing diversity across different knowledge types, or it can focus on questions where the model exhibits low confidence or high fluctuations during adaptation. Such sampling strategies based on training dynamics exist~\cite{swayamdipta2020dataset,pleiss2020identifying}, but they have not been applied to the task of zero-shot QA with KGs.  

\noindent \textit{RQ4: Do models generalize well to tasks with low domain overlap?} Models generally perform better on tasks that require knowledge similar to that in the training data, i.e., tasks with high domain overlap (HDO). This is confirmed by the relatively larger gains obtained when using ConceptNet for CSQA and ATOMIC for SocialIQA, compared to using these sources on datasets like WinoGrande~\cite{mitra2019exploring,Ma2021}. Whether models can generalize well to questions with low domain overlap (LDO) is an open question.

\noindent \textit{RQ5: What is the connection between model’s accuracy and properties of the task?} 
We analyze the impact of answer candidate similarity, answer candidate length, and vocabulary overlap on the model accuracy. We expect that the questions with different answer similarity and length would require different reasoning. For example, similar answer candidates would require models to focus on the relatively small differences between answers. Similarly, task partitions with higher vocabulary overlap to the synthetic data would intuitively be easier for the models. Prior work has reported that models rely on spurious correlations, such as lexical properties, to answer questions~\cite{gururangan2018annotation,mccoy2019right}.
Li et al.~\cite{li2021systematic} show that only considering the answer candidates may bring high performance on some tasks, indicating that the properties of the answer candidates have a large impact on the model. Similarly, fine-tuned models perform much better on questions that resemble the training data~\cite{ma2021exploring}. Yet, these investigations have not been conducted in a zero-shot setting.

\begin{figure*}[!t]
\centering
\includegraphics[width=0.75\linewidth]{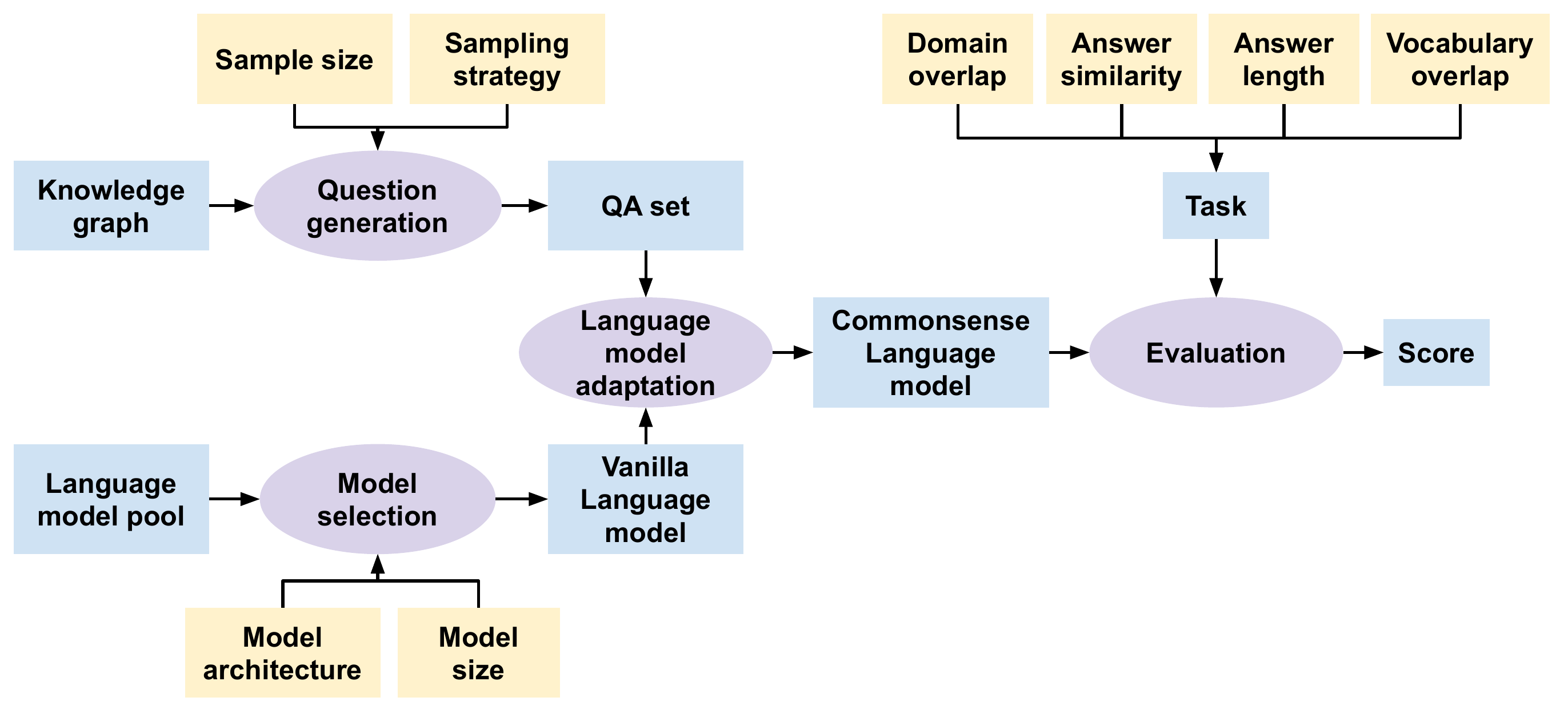}\\
\caption{Overview of our study. The question generation step takes knowledge graph as input, and yields a question answer set. The generated set depends on the sampling size and strategy. In parallel, a language model is chosen out of a pool of models based on two factors: architecture and size. The selected vanilla language model is adapted based on the synthetic QA set, resulting in a commonsense language model. The task used to evaluate this model depends on several factors: domain overlap, answer similarity, answer length, and vocabulary overlap. The combination of the choices for these four aspects determines the task, which is finally used to evaluate the commonsense model, and produce an accuracy score.}
\label{fig:overview}
\end{figure*}

\section{Method} 
\label{sec:method}

We follow the task formulation of \textit{generalizable commonsense reasoning} proposed by \citet{Ma2021}. The input consists of a natural language question $Q$ and $n$ candidate answers $A_i$, $|A_i|=n$. Exactly one of the candidate answers, marked with $A$, is correct. The remaining $(n-1)$ candidate answers serve as distractors. As we assume a zero-shot setup, the models have no access to the benchmark-specific training data. 
Each model is adapted once, after which they are fixed, and directly applied on test partitions of various benchmarks.

To address the task of generalizable commonsense reasoning, we assume a knowledge-driven QA framework, where pre-trained LMs are adapted with artificial QA sets derived from KG data. 
We create artificial QA sets with thousands of questions by sampling statements from the recently-introduced CommonSense Knowledge Graph (CSKG)~\cite{ilievski2021cskg}, and transforming them into multiple-choice questions.
Each question corresponds to a particular knowledge dimension (e.g., temporal or spatial knowledge)~\cite{ilievski2021dimensions}. We define \textit{domain} as the dimensions of common sense necessary for solving a particular set of tasks.
Given a natural language question $Q$, and $n$ candidate answers $\{A_{1}, ..., A_{n}\}$, the LM has to select the most probable answer $A$ during training. Once the LM adaptation is done, the updated LM is applied across QA tasks in a zero-shot manner. 

The setup of this study is visualized in Figure~\ref{fig:overview}. We investigate the performance of self-supervision of language models with knowledge graphs, in relation to: 1) size and architecture of the language model; 2) size and sampling strategies of the knowledge used for model adaptation; and 3) properties of the task, such as overlap with knowledge and answer length. We describe the language models, knowledge sampling, and task properties in this section.


\subsection{Language Models} 
\para{Model architectures}
We adopt two widely-used pre-trained models: RoBERTa~\cite{Liu2019RoBERTaAR} and  T5~\cite{raffel2019exploring}. RoBERTa is an encoder-only masked language model (MLM), whereas T5 is an encoder-decoder model which converts tasks into text-to-text format. 

Following~\citet{Ma2021}, for RoBERTa each input sequence is a concatenation of the question and one of its answer candidates. We mask one non-stop token in the sequence at a time, and compute the masked token's loss. We then take the averaged loss for the sequence and this is repeated for every answer candidates. We then train the model with the margin loss:
$$L=\frac{1}{n} \sum_{\mbox{\tiny$\begin{array}{c}
  i=1\\
  i\neq y\end{array}$}}^{n}max(0,\eta-S_y+S_i)$$ 
where $S_y$ and $S_i$ are the negative averaged loss for correct answer and distractor respectively. During inference, we take the candidate with highest score $S$ as the answer. 

For T5, each input sequence is the same as RoBERTa except that we add a task-specific prefix, ''reasoning:'', to the front following the adaptation of T5 to downstream tasks in the original paper~\cite{raffel2019exploring}. The model is pre-trained to generate "true" or "false" token. $L_{true}$ represents the loss of the "true" token logits, while $L_{false}$ represents the "false" token logits. For each answer candidate, we compute the score $S=L_{true}-L_{false}$ and use the same margin loss function as in RoBERTa to jointly predict the optimal candidate.\footnote{We also tried to score the answers individually, or to concatenate the question with all answer candidates, and teach the model to predict the position or make a copy of the right candidate, following \cite{khashabi2020unifiedqa}. These loss strategies performed consistently worse, and we leave them out of the paper.}



\para{Model sizes} We use RoBERTa's base and large models, which have 125M and 355M parameters, respectively. We experiment with three T5 models of different sizes: small (60M parameters), large (740M), and 3b (2.85B). More details about the model training can be found in the Appendix of this paper. 

\subsection{Knowledge sources and sampling}

We subselect knowledge from the CommonSense Knowledge Graph (CSKG)~\cite{ilievski2021cskg}, which combines seven commonsense knowledge resources under a shared representation, including ConceptNet~\cite{10.5555/3298023.3298212}, ATOMIC~\cite{DBLP:conf/aaai/SapBABLRRSC19}, and Visual Genome~\cite{krishna2017visual}. In total, CSKG contains over 7 million commonsense statements, which are used to describe over 2 million nodes with 58 properties. Each statement in CSKG consists of a head ($h$), relation ($r$), tail ($t$), and additional qualifiers ($q$).
Each knowledge statement in CSKG is categorized into one of the following 13 dimensions: lexical, similarity, distinctness, taxonomic, part-whole, creation, utility, comparative, quality, temporal, spatial, motivational, and relational-other~\cite{ilievski2021dimensions}.

\para{Sample sizes.} We use the subset of CSKG which combines ATOMIC, ConceptNet, WordNet~\cite{miller1995wordnet}, Wikidata~\cite{vrandevcic2014wikidata}, and Visual Genome. We use all relations in this subset. We sample QA sets that correspond to a percentage $K$ from this knowledge set. We experiment with $K \in  \{1, 5, 10, 33, 50, 100\}$. In comparison, Ma et al. \cite{Ma2021} use 100\% of the data for fourteen manually-selected semantic relations. 



\para{Sampling strategies.} The sampling of $K\%$ artificial examples brings up the question: how do we optimally select these questions? We experiment with seven selection strategies that are based on training indicators and the KG structure. 
(i) \textit{Random} draws $K\%$ of the question pool by chance, without replacement.
(ii) \textit{Dimension} selects the questions that belong to a knowledge dimension. Out of the thirteen dimensions defined in~\cite{ilievski2021dimensions}, we evaluate the five best populated dimensions: temporal, desire/goal, taxonomic, quality, and relational-other. Here we are interested to see if there is a particular knowledge dimension that is more beneficial than others. For fair comparison between sampling strategies, we limit the questions selected for a dimensions to the equivalent of $K\%$ of the entire question set.
(iii) \textit{Uniform} selects an equal number of questions from each of the thirteen dimensions. We limit the total number of questions to the equivalent of $K\%$ of the entire question set. 
(iv) \textit{Vanilla-confidence} selects questions based on the model confidence before any adaptation, i.e. we run the vanilla LM on the entire QA sets to get its confidence scores. We experiment with two variants, which select the questions with lowest and highest confidence, respectively. We would like to test if the model benefits more from questions it considers easy or hard. 
(v) \textit{Confidence} selects questions based on the mean model confidence for the true label across the adaptation epochs. Specifically, we first train a model on entire QA sets and record each questions' training statistics as in ~\cite{swayamdipta2020dataset}. Then we select the subsets of the data to train our actual task model. This method applies to the last two strategies as well. Here we use two variants: confidence-low and confidence-high, analogous to the vanilla-confidence strategy.
(vi) \textit{Variability} detects the $K\%$ of the questions with extreme standard deviation for the true label across the adaptation epochs~\cite{swayamdipta2020dataset}. We experiment with variability-low and variability-high sampling.
(vii) \textit{Margin} selects the $K\%$ with the most extreme mean difference between the confidence of the correct answer and the incorrect ones~\cite{pleiss2020identifying}. We consider margin-low and margin-high sampling.

For every strategy, $K=100$ corresponds to the entire synthetic QA pool, while $K=0$ is the vanilla pre-trained LM without adaptation. 


\subsection{Tasks} 

We evaluate on five benchmarks for multiple-choice commonsense question answering. \textit{CommonsenseQA (CSQA)}~\cite{talmor-etal-2019-commonsenseqa} is a five-choice question answering benchmark which evaluates a broad range of common sense aspects. \textit{SocialIQA (SIQA)}~\cite{sap-etal-2019-social} is a three-choice QA benchmark that requires reasoning about social interactions.
\textit{Abductive NLI (aNLI)}~\cite{bhagavatula2019abductive} is formalized as natural language inference, where, given the beginning and the ending of a story, the task is to choose the more plausible hypothesis out of two options.
\textit{PhysicalIQA (PIQA)}~\cite{Bisk2020} is a binary choice task, which tests the ability of models to employ physical reasoning.
\textit{WinoGrande (WG)}~\cite{sakaguchi2019winogrande} is a binary choice anaphora resolution task.
We measure the \textit{accuracy} of a language model on a benchmark as the ratio between the correctly-answered questions and the total number of questions.

\para{Task properties} We measure granular model performance by computing its accuracy on task partitions. We partition commonsense tasks based on four properties.

First, we consider model accuracy in relation to the \textit{domain overlap (DO)} between the KG and the task. Two of the five benchmarks are known to have high domain overlap (HDO) with existing KGs~\cite{mitra2019exploring,Ma2021}: CSQA has been devised based on knowledge in ConceptNet, while SocialIQA has been created based on the ATOMIC KG~\cite{DBLP:conf/aaai/SapBABLRRSC19}. The remaining three benchmarks have been created independently of the KGs, therefore, we consider them to have low domain overlap (LDO) with our KGs. We compare the model performance on the benchmarks with HDO and LDO.

Second, we partition the task into quartiles based on the \textit{answer similarity (AS)}  between the candidates. Here, we compute the answer similarity for a question $q$ through the Jaccard similarity between the tokens of the candidates $A_i$ and $A_j$: 
$
AS (q)=\frac{|T_{A_i}\cap T_{A_j}|}{|T_{A_i}\cup T_{A_j}|}
$. Here, 
$T_{A_i}$ and $T_{A_j}$ are the set of tokens of candidates $A_i$ and $A_j$, respectively. We limit the answer similarity analysis to the PIQA benchmark, which has two candidates for each question.

Third, we partition a task into quartiles based on the \textit{answer length (AL)} of its questions. We compute the answer length of a question $q$ by summing the tokens $T_{A_i}$ of the candidates $A_i$:
$
AL(q) = \sum_{i=1}^{n}{|T_{A_i}|}
$.

Fourth, we partition the task into quartiles based on the \textit{vocabulary overlap (VO)} between the task questions and the synthetic QA set. We first compute the frequency of every token that appears in the synthetic data. Given a task question, we compute the average frequency of the candidate tokens in the synthetic data.
To increase the effect of the tokens with low frequency, we use the reciprocal value of the token frequencies:
$VO(q)=\frac{1}{m} \sum_{
  k=1}^{m}\frac{1}{f(t_k)}$.
Here, $m$ is the number of tokens in the combination of the candidates ($|\bigcup_i^n{T_{A_i}}|$ for each question, $t_k$ is the $k$-th token in the answer candidates, and $f(t_k)$ is its frequency in the synthetic data. 

When splitting the task based on \textit{answer simillarity}, \textit{answer length}, and \textit{vocabulary overlap}, we use RoBERTa's tokenizer, and we focus on the PIQA benchmark which has a variety of properties among its questions.

\begin{table*}[!t]
	\centering
 	\small
	\caption{Evaluation results on five benchmarks of our five models with their optimal data size and sampling strategy. For comparison, we show results of relevant baselines. We show the results of the best model reported in each prior work. `*' indicates that the average is computed on an incomplete set of benchmarks, and is thus not directly comparable to the other results in the same column. Best results per column are shown in bold.}
	\label{tab:Results}
	\begin{tabular}{l  | rrr | rr | rrr}
	\hline
	    \multirow{2}*{\bf Model}&\multicolumn{3}{c|}{\bf LDO}& \multicolumn{2}{c|}{\bf HDO}&\multirow{2}*{\bf Avg(LDO)}&\multirow{2}*{\bf Avg(HDO)}&\multirow{2}*{\bf Avg}\\
	    &{\bf aNLI}&{\bf WG}&{\bf PIQA}&{\bf SIQA}&{\bf CSQA} & \\\hline
	    {\bf Majority~\cite{Ma2021}} & 50.8 & 50.4 & 50.5 & 33.6 & 20.9 & 50.6 & 27.25 & 41.2 \\
        {\bf RoBERTa-large~\cite{liu2019roberta}} & 65.5 & 57.5 & 67.6 & 47.3 & 45.0 & 63.5 & 46.1 & 56.6 \\ \hline
	     {\bf COMET~\cite{bosselut2019comet}}&-&-&-&50.1&-&-&*50.1&*50.1\\
	    {\bf Self-Talk~\cite{Shwartz2020UnsupervisedCQ}}&-&54.7&70.2&46.2&32.4&*62.5&39.3&50.9\\
	    {\bf SMLM~\cite{Banerjee2020SelfsupervisedKT}}&65.3&-&-&48.5&38.8&*65.3&43.7&50.9\\
	      {\bf Ma et al.~\cite{Ma2021}}&70.5 & 60.9 & 72.4 & 63.2 & 67.4 & 67.9 & 65.3 & 66.8\\ 
	      {\bf Dou \& Peng~\cite{dou2022zero}}&- & - & - & 59.9 & 67.4 & - & 63.6 & 63.6\\     \hline
        {\bf RoBERTa-base (ours)} & 59.9& 52.6& 65.3&54.4&53.4&59.3&53.9&57.1\\ 
	    \bf RoBERTa-large (ours)
        & 72.0& 60.2& 72.5&{\bf 65.4}&66.9&68.2&66.2&67.4\\ 
        \bf T5-small (ours) &50.6&52.2&56.0&42.5&36.9&52.9&39.7&47.6\\ 
		\bf T5-large (ours) &65.5&59.0&70.6&57.2&62.9&65.0&60.0&63.0\\ 
        \bf T5-3b (ours)
		&{\bf 76.6}&{\bf 71.0}&{\bf 76.7}&65.3&\bf 69.9&{\bf 74.7}&{\bf 67.6}&{\bf 71.9}\\ \hline
		\bf \textit{RoBERTa-large (supervised)} & 85.6 & 79.3 & 79.2 & 76.6 & 78.5 & 81.4 & 77.5 & 79.8  \\ \hline
	\end{tabular}
\end{table*}

\begin{figure}[!t]
\centering
\includegraphics[width=0.95\linewidth]{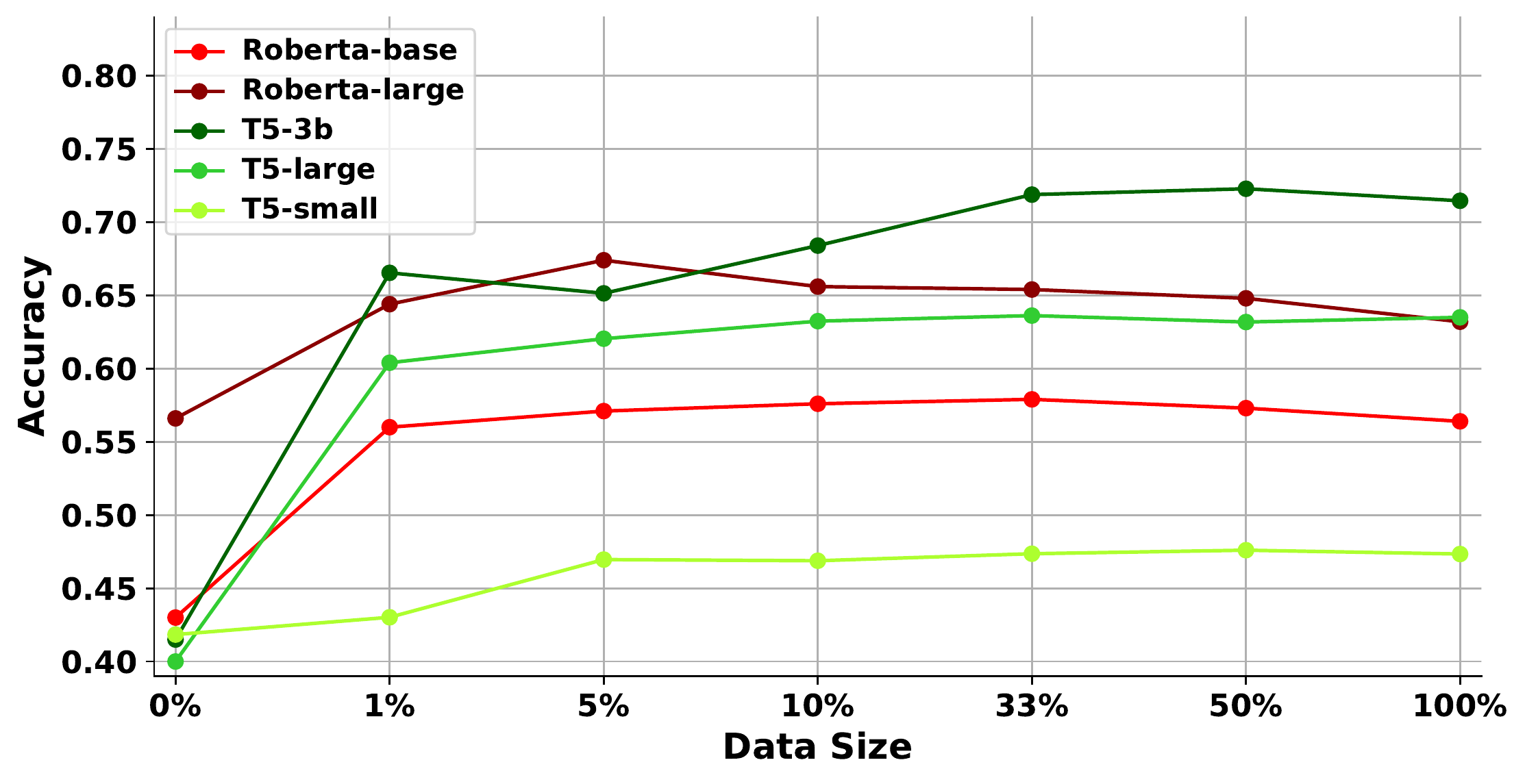}\\
\caption{Evaluation results of the five models with different data sizes, corresponding to 0, 1, 5, 10, 33, 50, and 100\% of the synthetic data. Each point represents the average performance of a model over the five benchmarks.} 
\label{fig:data_sizes}
\end{figure}

\section{Results}
\label{sec:results}

We provide results based on applying the methodology presented in Section~\ref{sec:method} to address the research questions in Section~\ref{sec:hypotheses}.

\noindent \textbf{What is the overall impact of model and knowledge choice on the generalizability of self-supervision of LMs with KGs (RQ1)?} Table~\ref{tab:Results} shows the results obtained with the best performing knowledge sampling strategy (random) and the best data size per model architecture (5\% for RoBERTa models, 33\% for T5 models). The results show that the best performance is clearly obtained with self-supervision of the model T5-3b with 33\% of the data. The zero-shot result of this model is 71.9 on average over the five benchmarks, which is 15.3 points higher than the vanilla RoBERTa-large model and 5.1 points higher than the previous state-of-the-art result of Ma et al. This result is especially encouraging in comparison with the supervised RoBERTa-large LM, which is now only 7.9 points higher than our result, despite relying on benchmark-specific training data. Our second best model is RoBERTa-large, which is able to outperform the RoBERTa-large model in Ma et al. despite relying on only 5\% of the training data. As expected, the accuracy of the Roberta and T5 models shows a clear positive impact of the model size, as T5-3b $>$ T5-large $>$ T5-small and RoBERTa-large $>$ RoBERTa-base.


\noindent \textbf{How much data is needed to adapt LMs to commonsense reasoning tasks (RQ2)?}
We investigate the impact of the synthetic data size on the five models (RoBERTa-base and large; T5-small, large, and 3b) in detail. Figure~\ref{fig:data_sizes} shows the average accuracy for the five models across the five benchmarks. We observe that models have different optima in terms of the data size that they are pre-trained with, which is largely determined by the model architecture. The encoder-only model, RoBERTa, performs best with about 5\% of the synthetic data. Specifically, RoBERTa-large achieves its optimum of 67.4\% accuracy with only 5\% of the data, whereas the performance of RoBERTa-base increases only marginally with more than 5\% of the data. 
Meanwhile, the generative encoder-decoder model, T5, benefits from more commonsense data. This is especially the case for T5-3b, our largest model, whose accuracy grows around 7\% by switching from 5 to 33\% of the data. Yet, the performance of the models generally peaks around 33\%, and plateaus with the increase of the data size. In our experiments, we use 5\% of the data for RoBERTa and 33\% for T5.

\begin{figure}[!t]
\centering
\includegraphics[width=0.95\linewidth]{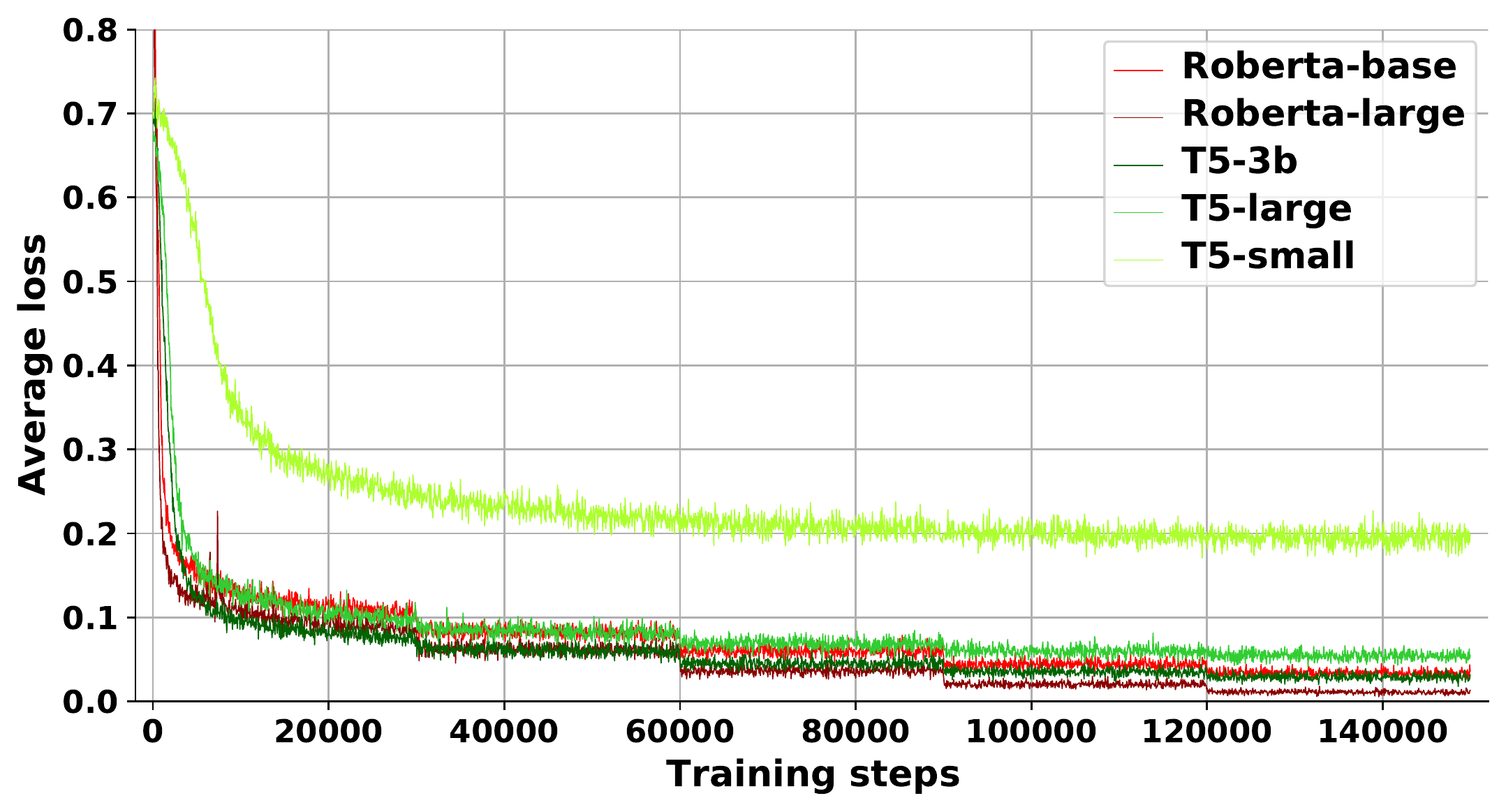}\\
\caption{Training curves of the models: RoBERTa-base, RoBERTa-large, T5-small, T5-large, and T5-3b. We use all of the models with 100\% of our training data}
\label{fig:training_curves}
\end{figure}

The need for more data of T5 models, as well as their ability to benefit less from it given same-sized LMs, can be attributed to their architecture. The learning curves in Figure~\ref{fig:training_curves} show that RoBERTa can adapt quicker to the task, which is because the adaptation loss corresponds to its pre-training loss. T5 starts with a high loss value because it has to learn the new prefix introduced in our work. This difference in the LM architecture also explains why RoBERTa-large outperforms T5-large, despite it being a smaller LM.

 

\begin{table*}[!t]
	\centering
	\small
	\caption{Evaluation results on the five benchmarks of {\bf RoBERTa-large} with different sampling strategies. All samples have equivalent sizes, corresponding to 5\% of the training data.  The best result per column is marked in bold.}
	\label{tab:sampling_roberta}
	\begin{tabular}{cc | rrr | rr | rrr}
	\hline
	    \multirow{2}*{\bf Strategy}&&\multicolumn{3}{c|}{\bf LDO}& \multicolumn{2}{c|}{\bf HDO}&\multirow{2}*{\bf Avg(LDO)}&\multirow{2}*{\bf Avg(HDO)}&\multirow{2}*{\bf Avg}\\
	    &&{\bf aNLI}&{\bf WG}&{\bf PIQA}&{\bf SIQA}&{\bf CSQA} & \\\hline
	    {\bf Random}&5\%&72.0&60.2&{\bf 72.5}&{\bf 65.4}&{\bf 66.9}&68.2&{\bf 66.2}&{\bf 67.4}\\\hline
        \multirow{5}*{\bf Dimension}&temporal&{\bf 72.7}&61.1&72.1&62.3&65.8&{\bf 68.6}& 64.1&66.8\\
        &desire&70.2&59.5&72.4&60.9&64.3&67.4&62.6&65.5\\
        &taxonomic&67.0&58.0&69.2&0.51&59.0&64.7&55.0&60.8\\
        &quality&71.3&{\bf 61.8}&72.0&58.5&64.6&68.4&61.6&65.6\\
        &rel-other&65.3&55.5&69.7&51.5&58.1&63.5&54.8&60.0\\\hline
        {\bf Uniform}&&69.6&58.0&72.4&61.7&64.3&66.7&63.0&65.2\\\hline
        \multirow{2}*{\bf Vanilla-conf}&high&63.3&59.1&67.6&49.4&47.2&63.3&48.3&57.3\\
        &low&57.9&51.9&55.6&33.1&21.7&55.1&27.4&44.0\\\hline
        \multirow{2}*{\bf Conf}&high&66.2&58.9&70.3&59.4&62.2&65.1&60.8&63.4\\
        &low&71.4&59.2&72.1&62.6&65.7&67.6&64.2&66.2\\\hline
        \multirow{2}*{\bf Varibility}&high&67.4&56.8&65.5&48.2&44.0&63.2&46.1&56.4\\
        &low&65.4&56.0&68.6&54.4&61.0&63.3&57.7&61.1\\\hline
        \multirow{2}*{\bf Margin}&high&67.1&58.2&70.7&60.1&62.3&65.3&61.2&63.7\\
        &low&72.3&60.5&71.2&62.7&65.0&68.0&63.9&66.3\\\hline

	\end{tabular}
\end{table*}

\begin{table*}[!ht]
	\centering
	\small
	\caption{Evaluation results on the five benchmarks of {\bf T5-large} with different sampling strategies. All samples have equivalent sizes, corresponding to 5\% of the training data. The best result per column is marked in bold.}
	\label{tab:sampling_t5}
	\begin{tabular}{cc | rrr | rr | rrr}
	\hline
	    \multirow{2}*{\bf Strategy}&&\multicolumn{3}{c|}{\bf LDO}& \multicolumn{2}{c|}{\bf HDO}&\multirow{2}*{\bf Avg(LDO)}&\multirow{2}*{\bf Avg(HDO)}&\multirow{2}*{\bf Avg}\\
	    &&{\bf aNLI}&{\bf WG}&{\bf PIQA}&{\bf SIQA}&{\bf CSQA} & \\\hline
        {\bf Random}&5\%&65.9&56.5& 70.5&55.4&61.9& 64.3&58.7&62.0\\\hline
        \multirow{5}*{\bf Dimension}&temporal& 66.6&56.4&{\bf 71.2}&54.9&{\bf 63.4}&{\bf 64.7}&{\bf 59.2}&{\bf 62.5}\\
        &desire&64.4&57.9&69.6&55.9&62.2&64.0&59.1&62.0\\
        &taxonomic&61.8&54.0&66.8&52.8&57.5&60.9&55.2&58.6\\
        &quality&{\bf 66.8}&{\bf 58.4}&70.0&{\bf 56.4}&59.6&65.1&58.0&62.2\\
        &rel-other&61.0&52.5&65.9&51.7&54.0&59.8&52.9&57.0\\\hline
        {\bf Uniform}&&65.3&57.5&69.2&56.6&62.7&64.0&59.7&62.3\\\hline
        \multirow{2}*{\bf Vanilla-conf}&high&65.3&56.8&69.0&55.5&57.5&63.7&56.5&60.8\\
        &low&64.0&56.0&68.1&52.0&59.6&62.7&55.8&59.9\\\hline
        \multirow{2}*{\bf Conf}&high&62.9&53.8&66.5&53.9&57.0&61.1&55.5&58.8\\
        &low&41.8&48.5&42.0&24.7&07.7&44.1&16.2&32.9\\\hline
        \multirow{2}*{\bf Varibility}&high&64&54.6&65.1&51.1&54.5&61.2&52.8&57.9\\
        &low&61.7&54.9&66.8&52.7&55.9&61.1&54.3&58.4\\\hline
        \multirow{2}*{\bf Margin}&high&63.8&54.5&67.2&52.8&56.9&61.8&54.9&59.0\\
        &low&41.5&45.0&43.7&24.1&09.1&43.4&16.6&32.7\\\hline
	\end{tabular}
\end{table*}

\begin{table}[!t]
\small
\caption{Examples of benchmark questions that are correctly answered with only one model, which is adapted with dimension-based knowledge. (*) denotes the correct answer.}
\begin{tabular}{l}
\tabucline[1.1pt]\\
dimension: temporal\\
Q:Jan went out with Quinn's friends and had a great time.\\What does Jan 
need to do before this?\\
A1:get dressed(*); A2:cancel her plans; A3:see Quinn's Friends again\\\tabucline[1.1pt]\\
dimension: desire\\
Q:Robert has no regret for punching Justin in the nose\\ because \_ was the
victim of injustice.\\
A1:Robert(*); A2:Justin\\\tabucline[1.1pt]\\
dimension: quality\\
Q:What can machines do that humans cannot?\\
A1:fail to work; A2:perform work; \\A3:answering questions; A4:see work\\
A5:fly(*)\\\tabucline[1.2pt]\\
\end{tabular}
\label{tab:examples}
\end{table}

\noindent \textbf{How to best sample questions for model adaptation (RQ3)?} We study the impact of different sampling strategies on RoBERTa-large and T5-large, given that they are in the same order of magnitude (hundreds of millions of parameters). We focus on 5\% of the data for computational reasons. The results for RoBERTa-large (Table \ref{tab:sampling_roberta}) and T5-large (Table \ref{tab:sampling_t5}) reveal that the random sampling strategy is surprisingly robust. Random sampling performs best for RoBERTa and it comes close to the best performing strategy (temporal) for T5-large. Besides random sampling, we observe consistently strong performance when training with some dimensions of knowledge: temporal, desire, and quality knowledge. Adapting the RoBERTa-large model with almost any dimension, except low vanilla confidence and high variability, performs better than the vanilla RoBERTa-large baseline.

The finding that random sampling leads to a strong and balanced model is consistent with the finding that random sampling of distractors is better than heuristic- and embedding-based strategies~\cite{Ma2021}. Both of these findings show that the natural distribution of the data provides diversity which is difficult to match with more focused strategies. Yet, the strong performance of the dimension-based strategies, whose data samples are disjoint by design, indicates that models trained with these dimensions capture complementary knowledge. This is confirmed in Table~\ref{tab:examples}, which shows three benchmark questions which are only answered correctly with one of the dimension-based models. The temporal model correctly solves the example that requires temporal ordering, while the desire/goal and quality models uniquely solve the questions that require reasoning about human psychology. This result shows that, while random sampling might be the optimal strategy to create a single model, multiple complementary models could be combined to achieve better accuracy.


\noindent \textbf{Do models generalize well to tasks with low domain overlap (R4)?}
Table~\ref{tab:Results} shows that the average improvement of T5-3b is mostly due to its improved performance on LDO benchmarks. T5-3b's improvement over RoBERTa is on average 6.5\% on the LDO benchmarks, but only 1.4\% on the HDO benchmarks. This generalization ability of T5-3b can largely be attributed to the larger capacity of T5-3b, which allows it to represent additional knowledge and associations between terms. In addition, this Table shows that the HDO benchmarks have been much more popular in prior work, and much larger gains over the vanilla LM have been reported on them (up to 15.1 points on SIQA and 22.4 points on CSQA). Conversely, results on the LDO benchmark  have rarely been reported in prior work on zero-shot commonsense reasoning, and the maximum improvement obtained in prior work is only 4.4 points on average across these benchmarks. Therefore, our accuracy improvement of 0.3 points for RoBERTa and 6.8 points for T5-3b is a notable leap towards robust performance on domains with low overlap.

\begin{figure*}[!t]
\centering
\includegraphics[width=\linewidth]{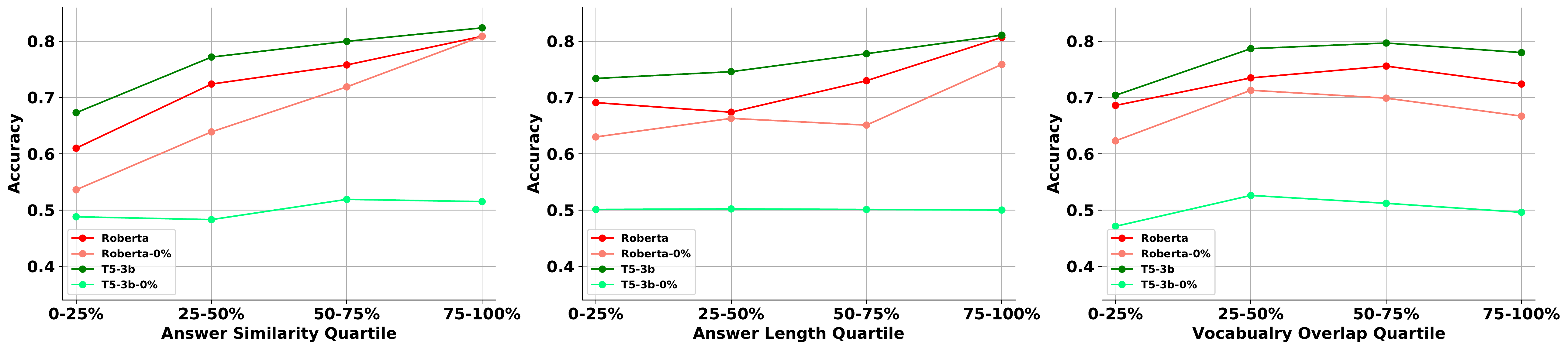}\\
\caption{Accuracy of the best performing RoBERTa-large and T5-3b models in relation to the answer similarity, answer length, and vocabulary overlap between the data used for pretraining and testing.}
\label{fig:analysis}
\end{figure*}

\begin{table*}[!t]
	\centering
 	\small
	\caption{Evaluation results on the similarity, length, and vocabulary overlap quartiles of PIQA data for the models RoBERTa  and T5-3b  with different data sizes. Best results per model and similarity quartile are marked in bold. }
	\label{tab:Data_size_analysis}
	\begin{tabular}{cc |rrrr|rrrr|rrrr}
	\hline
	    \multirow{2}*{\bf Model}&\multirow{2}*{\bf Data Size}&\multicolumn{4}{c|}{\bf Similarity}& \multicolumn{4}{c|}{\bf Length}&\multicolumn{4}{c}{\bf Vocabulary overlap}\\
	    &&{\bf 25\%}&{\bf 50\%}&{\bf 75\%}&{\bf 100\%}&{\bf 25\%}&{\bf 50\%}&{\bf 75\%}&{\bf 100\%}&{\bf 25\%}&{\bf 50\%}&{\bf 75\%}&{\bf 100\%}\\\hline
	    \multirow{6}*{\bf Roberta}
	    &0\%&53.6&63.9&71.9&\bf 80.9&63.0&66.3&65.1&75.9&62.3&71.3&69.9&66.7\\
	    &1\%&56.6&\bf 73.5&75.6&78.7&\bf 68.8&\bf 69.6&68.0&78.0&\bf 68.6&71.3&74.3&70.2\\
        &5\%&\bf 60.3&72.4&\bf 76.5&80.4&66.7&68.3&\bf 72.1&\bf 82.6&\bf 68.6&\bf 73.5&\bf 74.9&\bf 72.6\\
        &10\%&58.2&71.1&73.2&79.8&67.1&65.9&70.2&79.1&68.0&72.4&70.6&71.3\\
        &33\%&58.8&72.0&74.7&78.0&68.0&68.7&70.6&76.3&66.9&71.7&72.8&72.2\\
        &50\%&57.1&70.4&73.9&80.2&66.4&66.1&71.2&77.8&65.8&70.0&\bf 74.9&70.9\\
        &100\%&55.6&68.3&69.3&74.8&62.7&65.2&66.9&73.0&63.4&65.9&71.9&66.7\\\hline
        \multirow{6}*{\bf T5-3b}
        &0\%&48.8&48.3&51.9&51.5&50.1&50.2&50.1&50.0&47.1&52.6&51.2&49.6\\
        &1\%&61.7&73.9&73.6&78.7&71.2&70.0&70.2&76.5&68.2&70.4&76.9&72.4\\
        &5\%&60.3&72.4&71.9&79.3&68.4&68.7&67.8&79.1&64.1&70.7&77.6&71.7\\
        &10\%&65.4&75.9&75.4&82.6&70.6&74.8&72.5&81.3&69.9&74.1&77.8&77.4\\
        &33\%&67.3&\bf 77.2&80.0&82.4&\bf73.4&74.6&\bf 77.8&81.1&70.4&78.7& 79.7&78.0\\
        &50\% &\bf68.8&\bf77.2&\bf 80.2&\bf 84.3&73.2&\bf76.3&77.3&\bf83.7&72.1&\bf79.1&\bf80.2&\bf79.1\\
        &100\%&68.2&76.1&78.2&84.1&72.3&71.9&76.5&79.3&\bf78.9& 77.2&75.8&76.7\\
		\hline
	\end{tabular}
\end{table*}

\noindent \textbf{What is the connection between model’s accuracy and partitions of the task (R5)?}
Figure~\ref{fig:analysis} (left) shows that both RoBERTa-large and T5-3b perform better on questions with similar answers. Interestingly, vanilla RoBERTa already achieves high performance on this set, and pre-training only improves the performance on the questions with dissimilar answers. Given that the data used for pre-training is designed to only include questions with non-overlapping answers, this finding is intuitive, and explains the source of improvement of performance reported in prior work~\cite{Ma2021,dou2022zero}. T5-3b's accuracy gain over RoBERTa-large also owes to this property of the synthetic data, together with T5-3b's larger capacity to learn commonsense knowledge.
Moreover, Table~\ref{tab:Data_size_analysis} shows that both models perform better on the questions with dissimilar answers when they are trained with more data. The models perform optimal on the questions with similar answers with less data. This confirms our explanation that the knowledge used for pre-training directs the models towards better performance on the questions with dissimilar answers.

Both models perform best on questions with longer answers, while T5-3b is advantageous for short answers (Figure~\ref{fig:analysis}, middle). Notably, the synthetic QA data mostly consists of short answers, showing again that the performance gain of T5-3b owes to its capacity to extend original knowledge during the commonsense adaptation stage.
Furthermore, we see that T5 is able to exploit maximum amount of data for short answers (Table~\ref{tab:Data_size_analysis}), which is expected, given that most of the synthetic questions are relatively short. When it comes to longer answers, T5 performs best with less data, which indicates that the pre-training data has limited utility for this set of questions. Curiously, this pattern is not observed for RoBERTa - RoBERTa is unable to leverage more than 1\% of the data to improve its performance on the questions with short answers. We hypothesize that this is due to the limited model capacity of RoBERTa, causing limited ability to store additional knowledge from the synthetic data.

We do not see a clear correlation between vocabulary overlap and model accuracy, both overall (Figure \ref{fig:analysis} right), as well as across different data sizes (Table \ref{tab:Data_size_analysis}). Further experiments are needed to explain this finding.

\section{Discussion}

Our experiments show that the choice of LM size and architecture, as well as knowledge size and sampling strategy, affects the ability of models to answer commonsense questions across benchmarks. Encoder-decoder models benefit from more data to learn from, whereas sampling strategies that balance across different aspects yield best performance. The performance gain of knowledge-based pretraining is due to questions with short and dissimilar answers.
Next, we revisit three key assumptions of this study, and provide an alternative inspired by the results of our experiments.

\noindent \textbf{1. From a single model to mixture of models} Random sampling is the optimal single sampling strategy, as it balances well between the different properties of the data. More specialized strategies, e.g., focusing on knowledge dimensions, perform well on subsets of the task, but underperform on other subsets. This model specialization questions the assumption that a single zero-shot model is sufficient to perform optimally on different aspects of common sense, and suggests that model combinations, such as Mixture of models~\cite{gururangan2021demix}, might provide a more comprehensive and trustworthy commonsense model. This would entail, e.g., combining models from different dimensions, or models that capture complementary training dynamics.

\noindent \textbf{2. From implicit to explicit zero-shot commonsense reasoning} In our current framework, the rich and diverse commonsense knowledge is taught to an LM through a large set of QA pairs. Given the simplicity of these questions, an implicit assumption of this study is that LMs can reverse engineer these questions to learn commonsense knowledge implicitly, and apply this newly acquired knowledge on unforeseen benchmarks, whose surface properties may be different, but their underlying commonsense knowledge may be largely shared. While this is a reasonable assumption, our commonsense models are black boxes, and they do not provide an explicit justification for their decisions. 
A natural extension of this work is to devise \textit{explainable models}, i.e., models whose output includes the explicit reasoning steps associated with a predicted answer.

\noindent \textbf{3. From question answering to more realistic tasks} A key aspect of our zero-shot framework is its generalization across QA tasks. The gap between zero-shot and fine-tuning performance is closing down, which brings a natural question: are zero-shot models, adapted through an adaptation of neural techniques with background knowledge, able to generalize across tasks and domains, or have they merely learned how to answer questions convincingly? To address this question, we propose a shift towards more realistic tasks that rely on common sense, such as 
story understanding~\cite{kalyanpur2020braid}, dialogue modeling~\cite{ghosal2021cider}, and embodied QA~\cite{das2018embodied}. The rich prior work that focuses on these tasks has assumed the existence of benchmark-specific training data; zero-shot models have not been thoroughly explored. 

\section{Related Work}

\para{Generalizable Commonsense Reasoning.}
UNICORN~\cite{lourie2021unicorn} investigates continual learning of commonsense knowledge from multiple benchmarks, ultimately aiming to perform well on all of the benchmarks. This work demonstrates that LMs can learn from commonsense benchmarks effectively and efficiently, reaching relatively high accuracy based on little training examples. Prefix-tuning~\cite{li2021prefixtuning} adapts language models, by keeping the model parameters intact, but extending them with a small set of additional parameters tuned separately for each benchmark.
Rather than updating model parameters, Autoprompt~\cite{autoprompt:emnlp20} extends the model input with trigger tokens, which are updated during training.
Such efforts share our vision to develop models that can generalize to multiple commonsense benchmarks simultaneously. However, these works assume availability of training data, while we focus on zero-shot commonsense QA models.

\para{Zero-shot Commonsense Reasoning.} 
Zero-shot commonsense reasoning methods often elicit knowledge from pre-trained LMs, by using self-talk clarification prompts~\cite{Shwartz2020UnsupervisedCQ} or asking LMs to generate contrastive explanations~\cite{paranjape-etal-2021-prompting}. Models can be taught to answer questions by adapting to an external dataset~\cite{abdou-etal-2020-sensitivity}. We differ from these approaches because our models are based on knowledge available in large KGs, rather than mere distillation from language models. As shown in prior work~\cite{Ma2021}, KG-based approaches achieve superior performance compared to pure LM-based methods for zero-shot commonsense QA.

To use KGs for zero-shot pretraining and evaluation, \citet{Banerjee2020SelfsupervisedKT} pre-train an LM to perform knowledge completion, whereas \citet{bosselut2020dynamic} enhance the question based on knowledge completion models, and score an answer candidate in relation to the context, question, and generated knowledge.
Our work is based on the framework of~\citet{Ma2021}, which generates synthetic QA pairs from a consolidated KG to pre-train LMs.~\citet{Ma2021} investigate the impact of different loss functions and knowledge sources, showing that margin loss performs better than masked language modeling, and that more knowledge generally performs better, though this might change depending on the knowledge-task alignment. Follow-up work by~\citet{dou2022zero} extends the framework of~\citet{Ma2021}. with several data transformation methods, out of which measuring consistency between different prompt versions performs best. These data transformation efforts yielded notable performance gains, leading to new state-of-the-art results. Our work is orthogonal to these efforts, because we perform a systematic study of model size and architecture, knowledge sampling and size, and task properties, which have not been investigated in detail in prior work on zero-shot commonsense QA with knowledge graphs~\cite{Ma2021,dou2022zero}.



\para{Model Generalization and Data Selection.}
\citet{sen-saffari-2020-models} analyzed LM's ability to generalize across five different QA datasets. 
\citet{ma2021exploring} showed that models can have drastically different performances by fine-tuning on different subset of the data. 
\citet{swayamdipta2020dataset} proposed to select training instances based on models' confidence and variability, and they show that training on less-confident examples is more beneficial for generalization. 
Follow-up work~\cite{ethayarajh2021information} proposed an information-theoretic metric for estimating the difficult of a training example, treating as difficult the examples for which information is missing.
\citet{pleiss2020identifying} propose to identify erroneous data points based on their rank in the area under the margin of a machine learning model.
While prior work analyses model robustness by sub-sampling instances from the task's training set, we investigate the impact of knowledge selection, model selection, and task properties when models are adapted on large KGs for commonsense QA. \citet{ilievski2021dimensions} split synthetic data from KGs into 12 commonsense dimensions, revealing that some kinds of knowledge are much more useful for pre-training compared to others. Our study provides a comprehensive investigation of these prior efforts on the task of zero-shot commonsense QA with KGs. As such, prior work on subsetting training data based on knowledge dimensions or training indicators is integrated into our study.

\section{Conclusions}
\label{sec:conclusion}

This paper studied the impact of the knowledge, model, and task considerations on the self-supervision performance of LMs over KGs. We investigated different knowledge sizes and samples, as well as model architectures and sizes. We compared these against four task properties: domain overlap, answer similarity, answer length, and vocabulary overlap. We observed that the optimal knowledge size and sampling strategy is model-dependent, with encoder-only models learning quicker from less data than encoder-decoder models. Among the sampling strategies, random sampling performed best, closely followed with dimension-based sampling with temporal, quality, and desire/goal knowledge. Most of the improvement of the largest generative model comes from questions with short answers and dissimilar answer candidates, which is expected, given that the synthetic data generated from the KG has these properties. 

These findings point to three key directions for future work that uses self-supervision with large KGs to create generalizable commonsense reasoning agents. The abilities of complementary sampling strategies can be exploited by combining them into a single model. To enable robustness and trust, next-generation zero-shot models should be able to explain their reasoning explicitly. Finally, subsequent work should generalize the self-supervision KG-based method to more advanced tasks, such as story completion~\cite{kalyanpur2020braid} and embodied QA~\cite{das2018embodied}. All our code and data is available at \url{https://drive.google.com/file/d/1FE1_moXTK2Zpe4HQat78L6cXyrwtK4lO/view?usp=sharing}.
\bibliography{anthology,custom}
\bibliographystyle{acl_natbib}

\appendix
\newpage
\subsection*{Implementation}
\label{ssec:parameters}

For T5 training, we add the prefix ``reasoning:'' in front of every concatenation of question and answer, then ask the model to predict ``1'' for true, and ``2'' for false.\\
Regarding libraries, we used python 3.7.10, pytorch 1.9.0 and transformers 4.11.3.\\
Among all the training sets, we are using learning rate of $1e^{-5}$ , batch size of 32, weight decay 0.01, training epochs of 5, adam-epsilon of $1e^{-6}$, $\beta1=0.9,\beta2=0.98$, warm-up proportion of 0.05, margin of 1.0.\\
For CPUs, we used Intel(R) Xeon(R) Gold 5217 CPU @ 3.00GHz (32 CPUs, 8 cores per sockets, 263GB ram).\\
For GPUs, we used Nvidia Quadro RTX 8000, and Nvidia Geforce 2080Ti.

\end{document}